\documentclass[10pt,twocolumn,letterpaper]{article}

\usepackage{iccv}
\usepackage{times}
\usepackage{epsfig}
\usepackage{graphicx}
\usepackage{amsmath}
\usepackage{amssymb}

\usepackage{caption}
\usepackage{bm}
\usepackage{color}
\usepackage{subcaption}
\usepackage{graphicx}
\usepackage{comment}
\usepackage{dblfloatfix}

\usepackage[breaklinks=true,bookmarks=false]{hyperref}

\iccvfinalcopy 

\def\iccvPaperID{****} 

\begin{document}

\title{Video-to-Video Translation for Visual Speech Synthesis}

\author{Michail C. Doukas, Viktoriia Sharmanska, Stefanos Zafeiriou\\
Department of Computing, Imperial College London, UK\\
{\tt\small \{michail-christos.doukas16,sharmanska.v,s.zafeiriou\}@imperial.ac.uk}
}

\maketitle


\begin{abstract}
Despite remarkable success in image-to-image translation that celebrates the advancements of generative adversarial networks (GANs), very limited attempts are known for video domain translation. We study the task of video-to-video translation in the context of visual speech generation, where the goal is to transform an input video of any spoken word to an output video of a different word. This is a multi-domain translation, where each word forms a domain of videos uttering this word. Adaptation of the state-of-the-art image-to-image translation model (StarGAN) to this setting falls short with a large vocabulary size.  Instead we propose to use character encodings of the words and design a novel character-based GANs architecture for video-to-video translation called Visual Speech GAN (ViSpGAN). We are the first to demonstrate video-to-video translation with a vocabulary of 500 words. 
\end{abstract}
\section{Introduction}
Creating synthetic samples of a face displaying various expressions or uttering words and sentences is a very important problem in the intersection of computer vision, graphics and machine learning. The solutions were mainly given by the graphics community \cite{thies2016face, vdab} with a lot of manual work or by devising strictly person specific solutions \cite{obama}. Recently, with the advent of Deep Convolutional Neural Networks (DCNNs) and particularly with the introduction of Generative Adversarial Networks (GANs) \cite{gansoriginal} there has been a shift in focus. That is, instead of designing personalized methods and sophisticated rigging architectures, the focus is on designing machine learning architectures, based on GANs, which could harness the availability of data and simulate both, the physical process of facial motion creation and visualization. In particular, our paper focuses on a GANs-based model that can transform facial motion when uttering words. 

\begin{figure}[h]
\begin{center}
\includegraphics[width=1.0\linewidth]{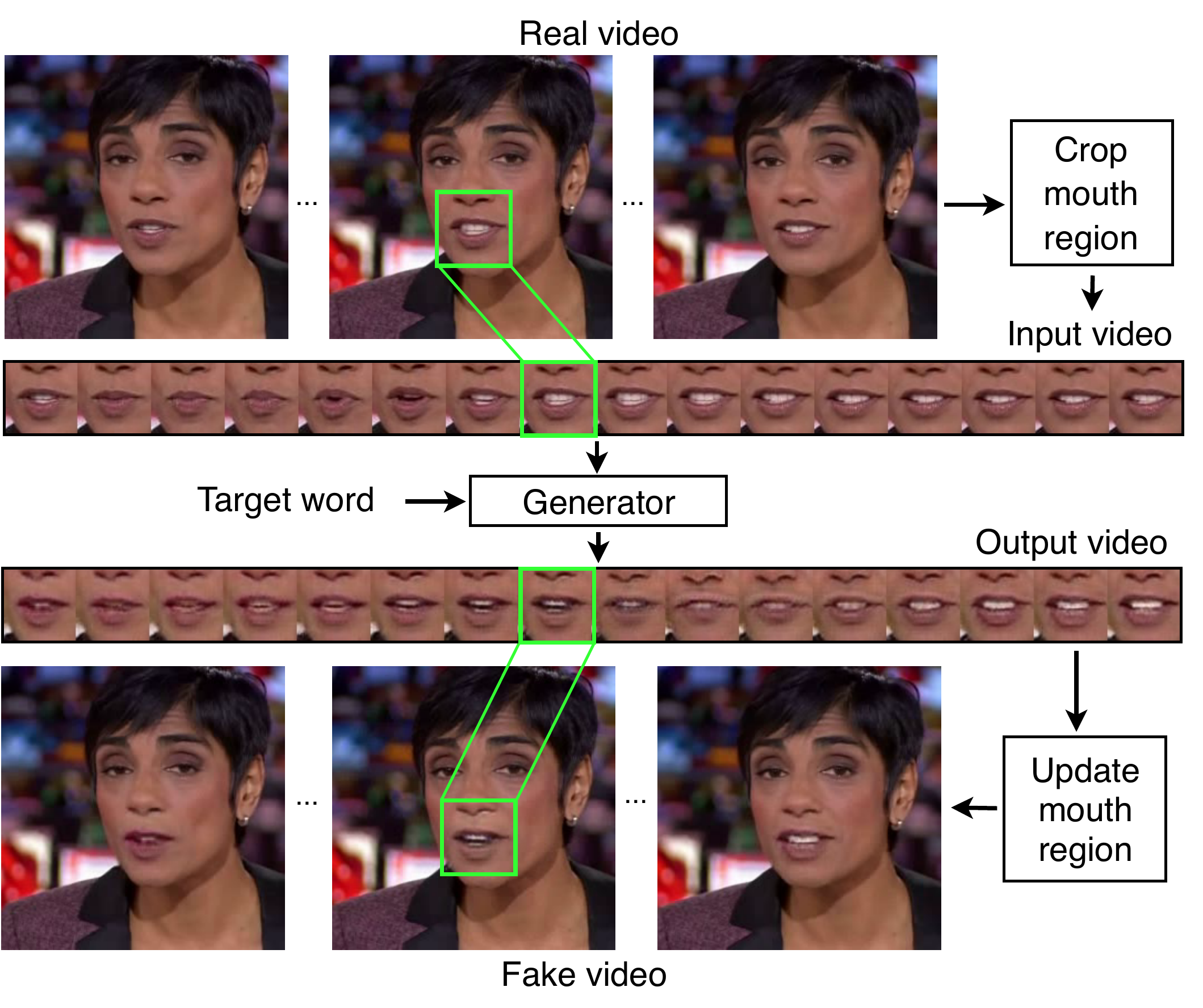}
\end{center}
   \caption{Video-to-video translation for visual speech synthesis: given a video of a speaker uttering \textit{"president"} and the target word \textit{"after"}, our generator produces a fake video of the speaker uttering the target word.}
\label{fig:fig1}
\end{figure}

In the seminal work\cite{gansoriginal}, GANs was introduced as an unsupervised machine learning method and was implemented by a system of two neural networks contesting with each other in a zero-sum game framework. 
Since then many different GANs architectures have been proposed, such as Deep Convolutional GANs (DCGAN) \cite{Radford2015UnsupervisedRL}, Wasserstein GANs (WGAN) \cite{WGAN}, Boundary Equilibrium Generative Adversarial Networks (BEGAN) \cite{Berthelot2017BEGANBE} and the Progressive GANs (PGAN) \cite{PGAN}, which was the first to show impressive results in generation of high-resolution images.

The concept of adversarial training has been also extended to the supervised learning settings such as dense regression (e.g., image-to-image translation) leading to various conditional GANs (cGAN) \cite{cgan} models \cite{pix2pix, CycleGAN2017, DiscoGAN, MultimodalUnsupervisedImagetoImageTranslation}. 
In image-to-image translation, the algorithms learn to synthesize a target image from a given input image, e.g. synthesizing photos from label maps, using a training set of aligned image pairs (e.g. pix2pix \cite{pix2pix}) which are difficult to collect. A recent work that made cGANs applicable in problems where paired trained data are not available is CyclicGANs \cite{CycleGAN2017} which used the notion of cyclic consistency. There is a plethora of works for the image-to-image translation task \cite{DiscoGAN, UNIT, TaigmanPW16, Bousmalis2017UnsupervisedPD, NIPS2017_6650, Shrivastava2017LearningFS, NIPS2016_6544}. 
A recent extension to a multi-domain setting has been also proposed in \cite{StarGAN}, where both data and labels are available but not paired.  
In order to support fully convolutional structures of the cGANs model \cite{StarGAN}, the authors append the one-hot label information as extra channels to the input image. The drawback of this strategy is that it is very difficult to train with many labels (domains). A similar approach was proposed in \cite{Ganimation} for transforming facial images depending to Action Units (AUs)\footnote{AU system is a systematic way to code the facial motion with respect to activation of facial muscles.}. 
Both state-of-the-art models, StarGAN \cite{StarGAN} and GANimation \cite{Ganimation}, allow transformations of the input face images with respect to hair color, age, gender, and anatomically-aware facial movements. 

Faces are one of the most widely used objects of choice for demonstrating the capabilities of GANs \cite{PGAN} and cGANs variants \cite{StarGAN, Ganimation, Zhou2018, UNIT, facetransfergans, DTGAN2018, Olszewski2017, TGANS2018}. This is the case because creating synthetic facial samples has countless applications, including video conferencing or movie dubbing for better lip-syncing results. 
%
In this paper, we aim at generating realistic synthetic videos of speakers uttering a given word. We assume that we are given a video of a person uttering one word along with a target word. Then, our method generates a new video with the same person uttering the target word. In particular, the contributions of our paper are:
\begin{itemize}
\item Our method extends the work of StarGAN \cite{StarGAN} to video-to-video translation using 3D convolutional structures.
\item In order to support many word labels, we propose to encode the words using character labels and design 
novel architectures for video translation, i.e. the generator network and the character inspector network. 
\item We are the first to show video-to-video translation with semantic transformation of 500 words/domains (using `in-the-wild' dataset LRW  \cite{Chung16}).
\end{itemize}

\section{Related work}

Unlike the image domain, very limited attempts have been successful in the task of unconditional video synthesis \cite{VondrickNIPS2016, tulyakov2017mocogan, Saito2017TemporalGA, RadfordMC15}. For the task of video style transfer, existing methods \cite{Chen_2017_ICCV, GuptaJAF17, 8100228, RDB18} focus on transferring the style of a given reference image to the generated video, however these methods cannot be directly applied to our problem of multi-domain video-to-video transfer. 
Conditional video generation have been primarily addressed with text inputs  \cite{VideoFromText2017, VideosfromCaptions2018}. 
To the best of our knowledge, there is only one prior work on general video-to-video synthesis \cite{wang2018vid2vid}. In this work, a mapping function is learned to translate a video input from one modality (such as segmentation maps) to a corresponding RGB video. Using aligned input and output video pairs, the proposed model learns to synthesize video frames guided by the ground truth optical flow signal. 
Hence, this method is not suitable for a video-to-video translation between semantically different domains, i.e. domains of different word utterings. 

The lipreading task is related to our problem, since we employ lipreading methods when forcing the generator to produce convincing videos. This task has been previously tackled using the word-level \cite{Chung16, StafylakisT17} and character-level \cite{Chung17} classification models. Visual speech synthesis is directly related to the objectives of our framework. This task has been primarily approached via pose transfer. That is, given a target speaker and a driving video, methods \cite{audioorvideo2video, thies2016face, Wiles18, facetransfergans, Olszewski2017, DTGAN2018, Zhou2018} generate video with visual speech via transferring the pose (lip movement) of one speaker to another. A different approach is speech-driven facial synthesis \cite{TGANS2018, Chung17b, audioorvideo2video, Wiles18, Zhou2018}. In this setting, the generated videos are conditioned on an image of the target speaker and an audio signal. 
The methods in \cite{TGANS2018} generate the facial animation of a person speaking providing only a facial image and a sound stream. Using a similar setting, the method of  \cite{Chung17b} produces video from audio, however it generates frames independently from each other and animates only the mouth region. In the recent work \cite{audioorvideo2video}, the authors use a GANs-based approach to generate talking faces conditioned on both, video and sound, which is suitable for audio-visual synchronization, and is different from our setup. 

\section{Method}
We use conditional adversarial neural networks to learn a mapping from an input video of a person uttering one word to an output video of the same person uttering a target word. The training is conditioned on the input video of a speaker, the label of the input word, and the label of the target word. Each word label corresponds to a video domain of speakers uttering this word. We consider a general setting of multi-domain translation, where any two words (in the vocabulary) can form source and target domains. In this setting, well-established approaches like pix2pix \cite{pix2pix}, CycleGAN \cite{CycleGAN2017} are not feasible, as they require training separate translation models for all possible pairs of labels. Instead we base our approach on the StarGAN model \cite{StarGAN} for multi-domain image-to-image translation. 

First, in Section \ref{subsection31}, we describe our method as an extension of the StarGAN framework for video-to-video translation, using 3D convolutions and words as class labels. 
In this approach, the word label is appended as extra channels to the input using one-hot representations, which is not scalable. 
Next, in Section \ref{subsection32} we describe our proposed framework \emph{Visual Speech GAN (ViSpGAN)} for video-to-video translation that uses character encoding of words for generation and could scale to any vocabulary size. 
%

\subsection{Word-level Video Translation} \label{subsection31}

We build upon StarGAN framework \cite{StarGAN} to form a generative model that synthesizes realistic videos of a speaker uttering words given an input video. Our model consists of three components: a conditional video generator $G$, a video discriminator $D$ and a video word classifier $D_{cls}$. 

Given an input video $\bm{X}$ with $T$ frames, $\bm{X} = \{\bm{x}_1, \bm{x}_2, \dots, \bm{x}_T\}$, and a target word label $\bm{l}_w$ as one-hot vector, the generator transforms $(\bm{X},\bm{l}_w)$ into a synthetic video $\bm{Y}$ of the same length, where the speaker from the input video $\bm{X}$ utters the target word $\bm{l}_w$. $(\bm{X},\bm{l}_w)$ denotes depth-wise concatenation of the video frames and the spatially replicated target label.
The discriminator $D$ is trained to distinguish real videos from the synthesized ones, while the word classifier $D_{cls}$ is trained to classify videos of uttered words based on word labels in the vocabulary. 
Therefore, the generator is forced to synthesize fake samples, which are both realistic -- similar to the ones in the dataset -- and represent the right target words. 

\paragraph{Adversarial Loss.}  
The objective of the conditional GAN-based model for synthesizing realistic videos can be expressed as a standard adversarial loss:
\begin{equation}
\nonumber
\mathcal{L}_{adv} = \mathbb{E}_{\bm{X}}[\log D(\bm{X})] + \mathbb{E}_{\bm{X}, \bm{l}_w}[\log (1-D(G(\bm{X}, \bm{l}_w))]. \\ 
\end{equation}
Here the generator $G(\bm{X}, \bm{l}_w)$ is conditioned on the input video $\bm{X}$ and the target label $\bm{l}_w$ and tries to minimize the adversarial loss, whereas the discriminator is conditioned only on the real video and aims to maximize this loss. 
Similarly to StarGAN, for training, we adopt the Wasserstein GAN objective \cite{pmlr-v70-arjovsky17a} with gradient penalty \cite{WGAN}:
\begin{equation}
\begin{split}
\mathcal{L}_{adv} = & \mathbb{E}_{\bm{X}}[D(\bm{X})] - \mathbb{E}_{\bm{X}, \bm{l}_w}[D(G(\bm{X}, \bm{l}_w))] \\ 
& - \lambda_{GP} \mathbb{E}_{\hat{\bm{X}}} [(|| \nabla_{\hat{\bm{X}}} D(\hat{\bm{X}}) ||_2-1)^2],
\end{split}
\label{eq:1}
\end{equation}
where $\hat{\bm{X}} = \alpha \bm{X} + (1- \alpha) \bm{Y}$  is sampled uniformly along a straight line between a pair of a real and a generated videos, $\alpha \sim \mathcal{U}(0, 1)$. 

\paragraph{Word Classification Loss.} 
The generator has to synthesize a video $\bm{Y}$ of uttering a target word $\bm{l}_w$. To ensure that the target word is uttered in the output video, an auxiliary word classifier $D_{cls}$ is introduced. It imposes the word classification loss when optimizing both the discriminator and the generator. 
By minimizing the the word classification loss on fake videos
\begin{equation}
\mathcal{L}^{fake}_{cls} = \mathbb{E}_{\bm{X}, \bm{l}_w}[-\log D_{cls}(\bm{l}_w|G(\bm{X}, \bm{l}_w))],
\label{eq:3}
\end{equation}
the generator learns to produce videos $G(\bm{X}, \bm{l}_w)$ with uttered target words $\bm{l}_w$ as seen by the classifier $D_{cls}$. 
On the contrary, by minimizing the corresponding classification term on real videos
\begin{equation}
\mathcal{L}^{real}_{cls} = \mathbb{E}_{\bm{X}, \bm{l'}_w}[-\log D_{cls}(\bm{l'}_w|\bm{X})],
\label{eq:2}
\end{equation}
the classifier $D_{cls}$ learns to predict correct word labels $\bm{l'}_w$ of the real videos. 
%
In this work, we employ the state-of-the-art lipreading neural network \cite{StafylakisT17} as a word classifier $D_{cls}$. 

\paragraph{Cycle Consistency Loss.} Adversarial training along with the word classification do not guarantee that the identity of the speaker of the input video is preserved in the generated video. 
Since we do not have a ground truth video $\bm{Y}$ available, we cannot compute a reconstruction error of the translation as in \cite{pix2pix}. Instead we apply a cycle consistency loss as in \cite{CycleGAN2017, StarGAN} to preserve the content of the input video through a cycle  application of $G$:
\begin{equation}
\mathcal{L}_{cyc} = \mathbb{E}_{\bm{X}, \bm{l}_w, \bm{l'}_w}[||G(G(\bm{X}, \bm{l}_w), \bm{l'}_w) - \bm{X}||_1].
\label{eq:4}
\end{equation}
Here the generator first synthesizes a speaker uttering the target word $\bm{l}_w$ and then translates this video $G(\bm{X}, \bm{l}_w)$ back to the same speaker uttering the input word $\bm{l'}_w$, for which we have the ground truth (input video $\bm{X}$).

The full objective for training the generator and discriminator networks of the 3D StarGAN model for video-to-video translation with word labels can be written as:
\begin{equation}
\begin{split}
&\mathcal{L}_G = \mathcal{L}_{adv} + \lambda_{cls} \mathcal{L}^{fake}_{cls} + \lambda_{cyc} \mathcal{L}_{cyc}\\
&\mathcal{L}_D = - \mathcal{L}_{adv} \\
&\mathcal{L}_{D_{cls}} = \mathcal{L}_{cls}^{real} 
\end{split}
\label{eq:5}
\end{equation}
where $\lambda_{cls}$ and $\lambda_{cyc}$ are hyper-parameters that control relative importance of the correspondent loss terms. 

In our implementation, we follow closely the network structures of the StarGAN model equipping them with 3D convolutions over video volumes. We refer to this model as 3D StarGAN\footnote{The only structural difference is that we have $D_{cls}$ and $D$ networks trained separately in 3D StarGAN, where as in StarGAN model for image translation, the classifier $D_{cls}$ is a part of the discriminator output $D$. This however did not work for video data.}.

\begin{figure}[h]
\begin{center}
\includegraphics[width=1.0\linewidth]{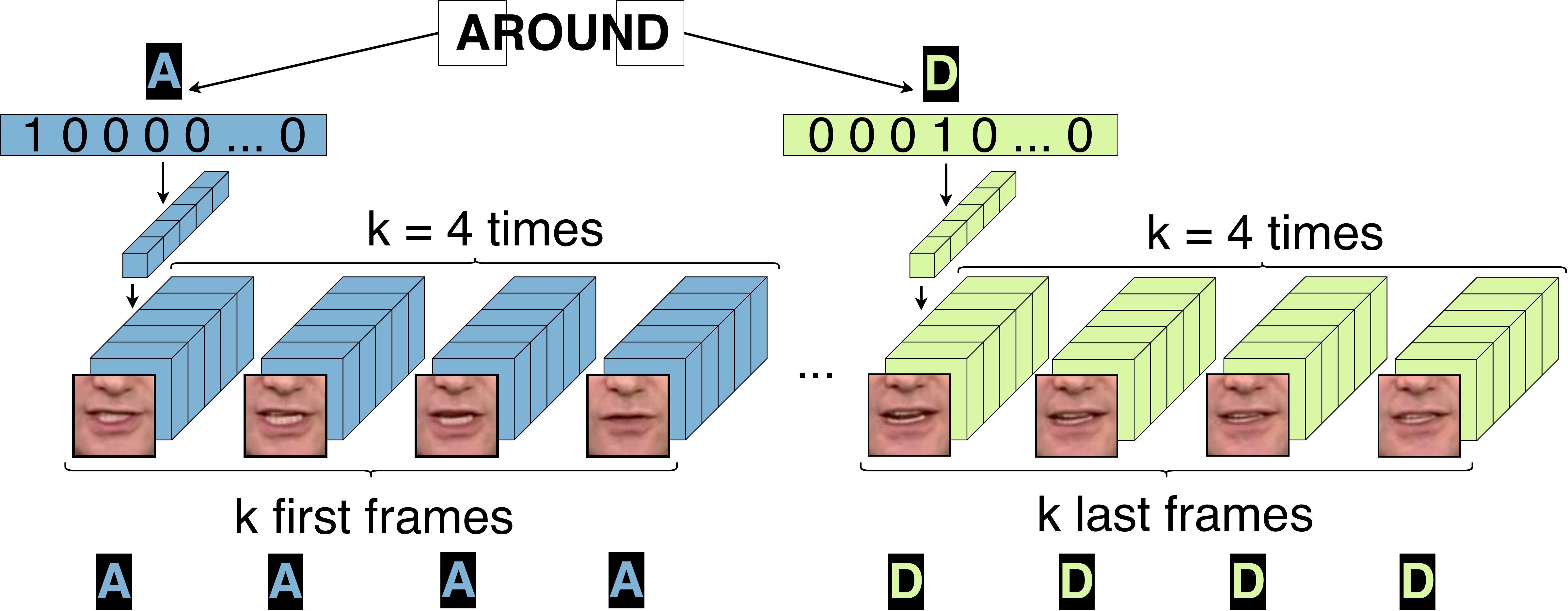}
\end{center}
   \caption{Illustration of the video generation with characters (as input to the generator network). Given the target word \emph{"around"}, and an input video of $T=24$ frames, each of the six characters in the word is replicated $4$ times (\emph{"aaaa-rrrr-oooo-uuuu-nnnn-dddd"}) to form labels for $24$ frames of the target video. After that, each character label ($26$-dimensional binary vector over English alphabet) is spatially replicated and then concatenated (as extra channels) to the corresponding frame of the input video using depth-wise concatenation.}
\label{fig:fig2}
\end{figure}

\begin{figure}[h]
\begin{center}
\includegraphics[width=1.0\linewidth]{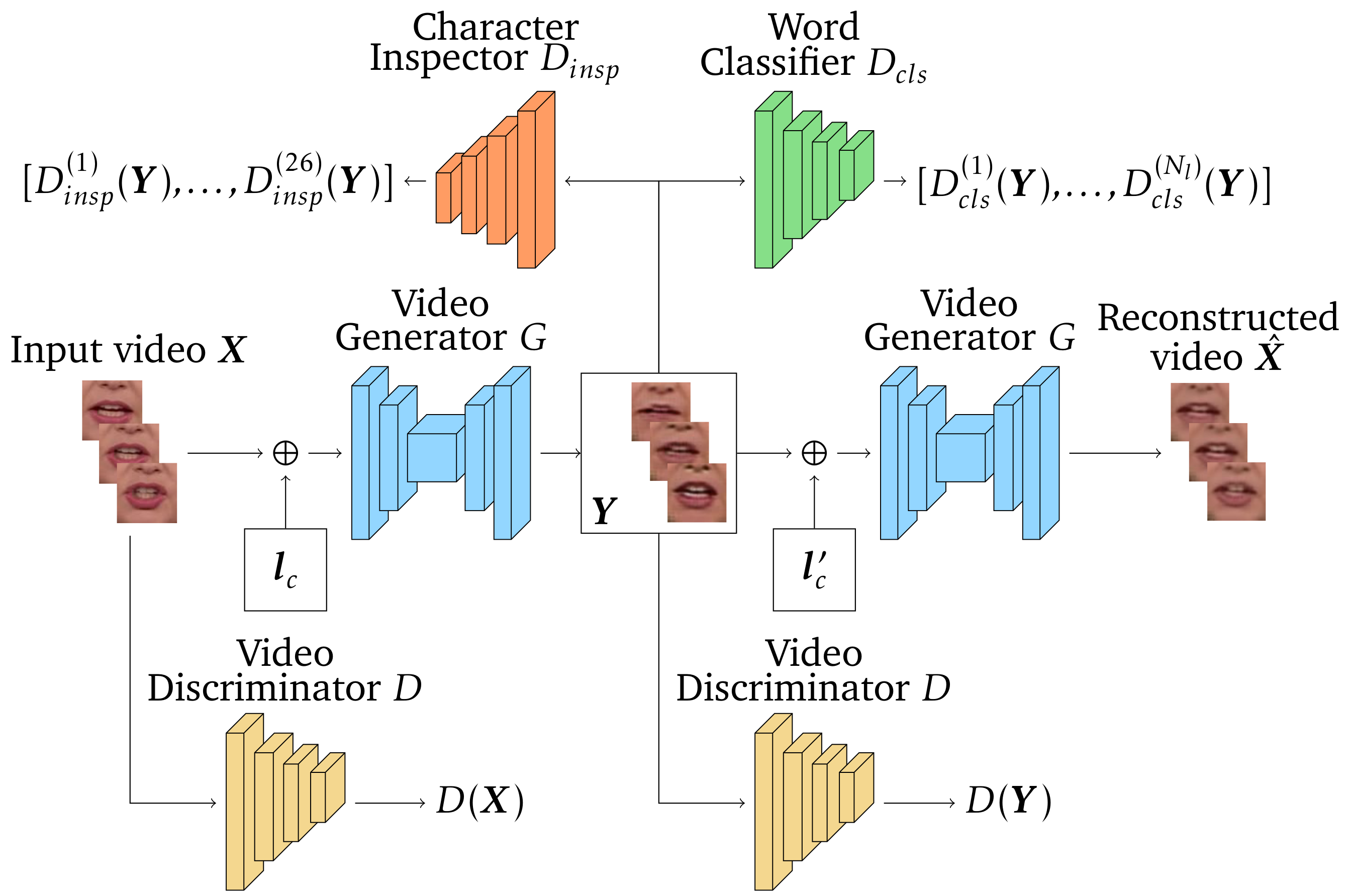}
\end{center}
   \caption{Overview of our proposed framework, which consists of four networks: the video generator $G$, the video discriminator $D$, the word classifier $D_{cl}$ and the character inspector $D_{insp}$. The generator synthesizes the fake video $Y$ conditioned on the input video and the target character labels and then maps $Y$ back to the original word domain again to get the reconstructed video $\hat{X}$. Additionally $D$ tries to distinguish between real or fake samples and $D_{cls}$ and $D_{insp}$ force $Y$ to contain the target word.}
\label{fig:fig3}
\end{figure}

\subsection{Character-level Video Translation} \label{subsection32}

Arguably the key design question of the conditional generative adversarial networks is how to incorporate the label information when training
the generator network. We found the idea of the StarGAN generator (to append the input RGB channels with the target labels) simple and remarkably effective. However, the 3D StarGAN extension for multi-word video translation has several shortcomings. First, it is difficult to train, in particular the generator network, even with a modest vocabulary size. Secondly, it does not account for commonalities in words and requires a lot of training data to learn all possible word translations. Finally, adding even one extra word to the vocabulary means expanding and retraining the generator from scratch. 

Instead we take a different approach and propose to use \emph{characters as multi-label embedding for word translations}. 
Our intuition is that similar words produce similar lip movements in the video frames where the common characters are uttered. For example, the words \textit{"because"} and \textit{"become"} share the same prefix \textit{"bec"}, which produces a similar lip movement for both words.
Character-level word embedding has several advantages for synthesis: (i) it can be scaled (potentially) to any vocabulary size, since we condition the generation on characters of the alphabet; (ii) it explores commonalities in words utterings, as words share characters. Finally, it can be naturally embedded in the functioning of the character inspector as will become clear shortly.


\paragraph{Video generation with characters.}
Throughout this work we consider uttering of English words, which can be generalized to other languages. The English alphabet has $26$ characters, and for simplicity we represent each character as a $26$-dimensional one-hot-vector\footnote{This can be further reduced to 5 dimensions with binary coding}.  
%
Given the target word $\bm{l}_w= \{\bm{c}_1, \bm{c}_2, \dots, \bm{c}_{N} \}$ with $N$ characters, we replicate each character $k = T/N$ times, to create a sequence of labels of the same length as the number of frames $T$: 
\begin{equation}
\bm{l}_c = \{\rlap{$\underbrace{\phantom{\bm{c}_1, \dots, \bm{c}_1}}_{k \text{ times}}$}\bm{c}_1, \dots, \bm{c}_1, \rlap{$\underbrace{\phantom{\bm{c}_1, \dots, \bm{c}_1}}_{k \text{ times}}$} \bm{c}_2, \dots, \bm{c}_2, \dots, \rlap{$\underbrace{\phantom{\bm{c}_{N}, \dots, \bm{c}_{N}}}_{k \text{ times}}$} \bm{c}_{N}, \dots, \bm{c}_{N}\}.
\label{eq:6}
\end{equation}
To form an input to the generator network, $(\bm{X}, \bm{l}_c)$, each character label is first spatially replicated and then concatenated (as extra channels) to the corresponding video frame using depth-wise concatenation. Please, see Figure \ref{fig:fig2} for an illustration. 

This simple strategy of conditional generation might look like each letter takes up the same time in the produced video. However, this is not the case, as the produced video uttering has to be realistic, and this is controlled by the discriminator networks. 
Next we describe how we enhance the GANs discriminator with a character-level classifier. \smallskip

\paragraph{Character Inspector.} 
%
The task of classifying videos of word utterings based on character sequences is not novel. A recently proposed model \cite{Chung17} addresses it by including an LSTM decoder to predict a sequence of distributions over character tokens and an attention mechanism. 
Since we do not aim to solve a lipreading task by itself, we adopt a simple yet effective approach. 
Instead of designing a character classifier that tries to predict the exact character sequence $\{c_1, c_2, \dots, c_{N} \}$ in the word, our approach involves \emph{inspecting} whether the characters are uttered in the video or not. We call this classifier network character inspector. 

Given a video $\bm{X}$ with a word label $\bm{l'}_w$, the character inspector $D_{insp}$ outputs a 26-dimensional vector $D_{insp}(\bm{X}) \in [0,1]^{26}$ with the probability for each of the characters being present in the word uttered in the video. 
In order to learn the task of inspection, we minimize the cross entropy loss between $D_{insp}(\bm{X})$ and the character-level embedding of the word, i.e. a $26$-dimensional binary vector $\bm{z'}$, where $\bm{z'}_i=1$ if the i-th character appears in $\bm{l'}_w$ and $\bm{z'}_i=0$ otherwise: 
\begin{equation}
\begin{split}
\mathcal{L}_{insp}^{real} &= \mathbb{E}_{\bm{X}, \bm{l'}_w}[\sum_{i=1}^{26} - {z'}_i \log D_{insp}^{i}(\bm{X}) \\
& - (1-{z'}_i) \log(1 - D_{insp}^{i}(\bm{X}))].
\end{split}
\label{eq:7}
\end{equation}
By minimizing $\mathcal{L}_{insp}^{real}$ on real videos from the dataset, $D_{insp}$ learns to recognize which letters are present/not present in the utterings. On the other hand, we utilize the same network to force the generator to synthesize videos with characters of the target word $\bm{l}_w$ by minimizing: 
\begin{equation}
\begin{split}
\mathcal{L}_{insp}^{fake} &= \mathbb{E}_{\bm{X}, \bm{l}_w}[\sum_{i=1}^{26} - z_i \log D_{insp}^{i}(G(\bm{X}, \bm{l}_c)) \\
& - (1-z_i) \log(1 - D_{insp}^{i}(G(\bm{X}, \bm{l}_c)))],
\end{split}
\label{eq:8}
\end{equation}
where $\bm{l}_c$ encodes the word label $\bm{l}_w$ (\ref{eq:6}), and cross entropy is computed between the predicted distribution $D_{insp}(G(\bm{X}, \bm{l}_c))$ and the character indicator $z_i$ of the target word $\bm{l}_w$, i.e.  $z_i = 1$ if character $i$ appears in $\bm{l}_w$ and $z_i = 0$ otherwise.

The advantage of using character inspector (as apposed to character detector, for example) is that it enables the GANs model to learn the \emph{appearance and duration of the characters} from the real videos and to reflect this knowledge when synthesizing fake videos. Not enforcing detection of the exact character sequence as conditioned in the video generation has also a shortcoming.  
%
The character inspector is unaware of the relative order of characters in the word or how many times they appear in the video. 
This could be addressed by incorporating an LSTM transducer to decode the exact letter sequence (which is an inherently harder problem than determining if a character is present in the video or not). Instead we combine the character inspector with the word classifier that clears away the need to learn the exact order of letters in the word. 

\paragraph{Feature Matching Loss.} Finally, we further enhance the lip movement features of the generated samples by adding a feature matching loss term to the objective of the generator, which is similar to the commonly used VGG loss \cite{wang2017highres, wang2018vid2vid}. The idea behind this loss is that two videos $\bm{Y}$ and $\bm{Y}^{*}$ of the same word uttering $\bm{l}_w$, should yield similar visual features when they pass through a pre-trained lipreading network. We use the LSTM of the word classifier network $D_{cls}$, and its output at the last time step $T$ as the visual feature representation $\bm{h}_T$ ($\bm{Y}$) of the video $\bm{Y}$. During training we sample a random video $\bm{Y}^{*}$ belonging to the target class $\bm{l}_w$ and we compute the $L_1$ distance between the visual features it yields and the visual feature vector which is obtained for the fake one $\bm{Y} = G(\bm{X}, \bm{\bm{l}_c})$. Therefore, the feature matching loss is defined as
\begin{equation}
\mathcal{L}_{fm} = \mathbb{E}_{\bm{X}, \bm{l}_w, \bm{Y}^{*}}[||\bm{h}_T(\bm{Y}^{*}) - \bm{h}_T(G(\bm{X}, \bm{l}_c))||_1],
\label{eq:9}
\end{equation}
and by minimizing $\mathcal{L}_{fm}$ with respect to $G$ parameters the generator learns to compose videos with visual features that match those of the real videos in the dataset. 

\paragraph{Visual Speech GAN Objective.} By combining all the loss terms, the generator of the proposed ViSpGAN model is trained under the objective function:
\begin{equation}
\begin{split}
\mathcal{L}_G = \mathcal{L}_{adv} &+ \lambda_{cls} \mathcal{L}^{fake}_{cls} + \lambda_{cyc} \mathcal{L}_{cyc} \\
&+ \lambda_{insp} \mathcal{L}^{fake}_{insp} + \lambda_{fm} \mathcal{L}_{fm}.
\end{split}
\label{eq:10}
\end{equation}
The discriminator $D$, the word classifier $D_{cls}$ and the character inspector $D_{insp}$ of ViSpGAN are optimized with respect to the losses
\begin{equation}
\begin{split}
&\mathcal{L}_D = - \mathcal{L}_{adv} \\
&\mathcal{L}_{D_{cls}} = \mathcal{L}_{cls}^{real} \\
&\mathcal{L}_{D_{insp}} = \mathcal{L}_{insp}^{real}.
\end{split}
\label{eq:11}
\end{equation}
The hyper-parameters are used to balance the contributions of the corresponding loss terms in comparison with $\mathcal{L}_{adv}$, and we set $\lambda_{cls}=1, \lambda_{insp}=1, \lambda_{cyc}=50$ and $\lambda_{fm}=50$. 

\subsection{Network Architecture}\label{subsection33}

\noindent \textbf{Video Generator $G$.} Based on \cite{StarGAN}, the generator is made up of three convolutional layers for downsampling the input, six residual blocks and another three transposed convolutional layer for upsampling. In order to adapt $G$ on the task of video generation, we replace 2D convolutions with spatio-temporal 3D convolutions. We note that downsampling is applied not only in the spatial but in the temporal dimension as well, reducing the dimension by a factor of two each time. In each layer, convolutions are followed by instance normalization \cite{InstanceNorm} and the ReLU activation function, except from the last one, where we simply apply a hyperbolic tangent function to obtain a confined output value for each pixel of the output video.\smallskip

\noindent \textbf{Video Discriminator $D$ and Character Inspector $D_{insp}$.} We adapt the PatchGAN \cite{DBLP:journals/corr/LiW16b, CycleGAN2017, pix2pix} setting for $D$, which identifies whether small video patches are real or fake, thus the output of the discriminator is not a single value, but a 3D volume. The discriminator receives a video of $T$ frames and uses a series of six 3D convolutions, which reduces the spatial dimensions by a factor of two, except from the last layer. On the other hand, the temporal dimension is being reduced only after the third layer, since it is much smaller than the spatial. In $D$ we do not apply instance normalization but we use Leaky ReLU with a leakage of 0.05. The same architecture is used for the character inspector as well, except from the final layer, where the spatio-temporal dimensions become unitary dimensions after downsampling and the feature dimension contains 26 values, each one indicating the presence or absence of the corresponding letter in the input video.\smallskip

\noindent \textbf{Word classifier $D_{cls}$.} This network is adopted from the work of \cite{StafylakisT17} on the task of lipreading. First, the input video is passed through a 3D convolutional layer, followed by instance normalization and ReLU. Next, we unstuck this 4D tensor (time $\times$ height $\times$ width $\times$ channels) in the temporal dimension, to get one 3D feature map for each time-step. Each one of these $T$ feature maps is passed through the same 34-layer ResNet \cite{He2016IdentityMI} for downsampling and then through an two-layered LSTM network. Finally, the last hidden state of the LSMT is mapped through a fully connected layer and a Softmax unit to a probability vector, with an equal length to the number of word classes $N_l$. 


\section{Experiments}
%
For empirical evaluations, we use the LRW dataset \cite{Chung16}, which is an `in the wild' audio-visual database of speakers uttering 500 different words. This database has been extracted from the BBC TV broadcasts and contains various speakers in many different poses. There are roughly 1000 short videos for each word class in the training set (500K videos), while the validation and test sets both contain 50 video samples per word class. Each clip consists of 29 frames and each frame is an $256 \times 256$ image. 
In order to overcome high computational demands when processing video inputs we mainly focus on the mouth area of the speakers. We extract facial landmarks from each frame of every video in the dataset using \cite{shen2015first}. Subsequently, we use the coordinates that correspond to the landmark in the center of the speaker's mouth to extract the region of interest, which is a cropped image of size $64 \times 64$. Since the generator performs down-sampling also in the temporal dimension, we choose to do cropping in the temporal dimension as well, such that the number of frames $T$ can be divided by 2 multiple times. In our experiments, we use $T = 24$, by simply dropping the first two and the last three frames in the video. The word is typically contained in the center of the video, and the first and last frames often include parts of the previous/next words from the sentence the video was extracted when the LRW was created. We use data augmentation by performing horizontal flips in the frames.
Finally, we divide the training set in two equal parts. The first spit is used to train the models and the second split is used to train an auxiliary network for the quantitative evaluations of the models, as it is further explained in Section \ref{subsection43}.


We conduct two sets of experiments. In the first one, we start with an ablation study of the ViSpGAN model and validate the individual effectiveness of 1) the word classifier $D_{cls}$, 2) the character inspector $D_{insp}$ and 3) the feature matching loss $\mathcal{L}_{fm}$ using a subset of 50 words from the LRW dataset. Then we provide a quantitative comparison of the proposed ViSpGAN model and the baseline method 3D StarGAN.  
%
In the second experiment, we qualitatively compare the generative performance of the models when trained on 50 words and report qualitative analysis of the ViSpGAN model when trained using the whole dataset of 500 words. Our implementation is based on TensorFlow and training the model on the entire dataset requires almost seven days on NVIDIA Tesla K80 GPU.

\subsection{Ablation Study}\label{subsection43}
\noindent \textbf{Setup.} We conduct this experiment using $50$ words. We arrange words in an alphabetical order and use all video samples from the first 50 word classes ($N_l = 50$). We train all models on the first split of the training set using the Adam optimizer with $\beta_1 = 0.5$, $\beta_2 = 0.999$, learning rate $\eta = 10^{-4}$ and a batch size of 8. A single generator update is performed after 5 discriminator and character inspector updates. We train all models for 20 epochs, with a linearly decreasing learning rate after the $10^{th}$ epoch, is the same manner as \cite{StarGAN}. We pre-train the word classifier $D_{cls}$, which remains fixed in all models. This classifier is trained on the first split of the training set, before the adversarial training takes place. For that, we use Adam with $\beta_1 = 0.5$, $\beta_2 = 0.999$, while exponentially decreasing the initial learning rate $\eta = 10^{-4}$ in discrete steps after each epoch. In addition, $L_2$ regularization is applied on weights with $\lambda = 0.0002$. We use a batch size of 16. 
For all our experiments, we produce fake samples using video inputs from the test set of LRW, since these videos have not been seen by the models during adversarial training. 

\paragraph{Quantitative evaluation.} 
To assess the performance of the proposed framework and all its variants in the ablation study, we test how well an independently trained auxiliary lipreading network could classify the fake samples generated by the models. For this reason we train an auxiliary classifier for the task of word recognition, using the second split of the training set, which has not been seen by our model during adversarial training. 
The architecture of this network is inspired by that of the word classifier $D_{cls}$, but it utilizes a ResNet-18 and a bi-directional GRU in the back end. After training this classifier for the first 50 words of LRW, we get an accuracy of 73\% on the test set. This performance is slightly inferior to state-of-art lipreading classifiers \cite{StafylakisT17, Chung17} on the same dataset, and can be improved by using a more sophisticated training strategy. We deliberately avoid reusing the architecture of the word classifier $D_{cls}$ here to be \emph{on a safe side}, i.e. the GAN models in the ablation study that have been trained by optimizing the word classification loss, have learned to `fool' the architecture of $D_{cls}$ by producing samples that it classifies correctly. 

\noindent In the ablation study, we compare the performance of the auxiliary classifier on synthetic samples produced by the ViSpGAN model and its variations: ViSpGAN without word classifier $D_{cls}$, ViSpGAN without feature matching $\mathcal{L}_{fm}$ or without both, ViSpGAN without character inspector $D_{insp}$. For each model, we generate $50$ fake samples per each of the $50$ target words using the unseen test set, so $2500$ test videos in total. We compute two metrics, the accuracy of the auxiliary classifier and the visual quality of the videos using the Fr{\'e}chet Inception Distance (FID) \cite{fid}. For the latter one, we use the auxiliary classifier (the last GRU state) as a feature extractor on real and fake videos. We compute the FID score using the real test videos and the fake videos produced by the generators of the models. We report the accuracy scores together with the FID scores in Table\ref{table:table1}.
\begin{table}[t]
\begin{center}
\begin{tabular}{|l|c|l|}
\hline
Model variation & Accuracy $\uparrow$ & FID $\downarrow$\\
\hline\hline
ViSpGAN w/o $\mathcal{L}_{fm}$, $D_{cls}$ & 75.5\% & $3.60 \pm 0.22$ \\
ViSpGAN w/o $\mathcal{L}_{fm}$, $D_{insp}$ & 89.2\% & $2.47 \pm 0.14$ \\
ViSpGAN w/o $\mathcal{L}_{fm}$ & 94.8\% & $2.79 \pm 0.11$\\  
ViSpGAN w/o $D_{cls}$ & 98.5\% & $4.11 \pm 0.20$\\ 
ViSpGAN & 99.1\% & $4.46 \pm 0.17$\\
3D StarGAN & 89.2\% & $2.80 \pm 0.19$ \\
\hline
ViSpGAN (500 words) & 95.6\% & $2.05 \pm 0.04$\\
\hline
\end{tabular}
\end{center}
\caption{Classification accuracy of the auxiliary word classifier on the videos generated by the model variations (the higher the better) and the FID scores (the lower the better).}
\label{table:table1}
\end{table}

As can be seen from Table \ref{table:table1}, the result on samples from the full model (ViSpGAN) enjoys the best performance comparing to the other variants. 
It is also clear that the the character inspector alone is not enough for synthesizing utterings of word structures, i.e. the auxiliary classifier has low accuracy on samples from the ViSpGAN w/o $\mathcal{L}_{fm}$, $D_{cls}$ model. 
In general, the auxiliary classifier has higher accuracy on samples generated by the models with the character inspector (ViSpGAN, ViSpGAN w/o $\mathcal{L}_{fm}$, ViSpGAN w/o $D_{cls}$). 
We conjecture that those generated videos contain word utterings which are easier to classify due to their more pronounced character presence. 
The feature matching component $\mathcal{L}_{fm}$ helps to generate more consistent visual features for each word class. This significantly increases the classification accuracy on fake videos and produces more pronounced lip movements. However, it comes at cost of the visual quality of samples, as can be seen by increased FID scores. 
The models without feature matching but with the word classifier (ViSpGAN w/o $\mathcal{L}_{fm}$) produce samples with better FID score but lower accuracy as evaluated by the auxiliary classifier.

Next, we compare the ViSpGAN model with our proposed adaptation of the StarGAN baseline model for video data, 3D StarGAN, when trained on $50$ words. We report the accuracy and the FID score in Table \ref{table:table1}. 
As can be seen from the results, the performance on samples from the 3D StarGAN is inferior to the proposed ViSpGAN model in terms of accuracy, but has better FID score. We also trained a variant of the ViSpGAN model with the word classifier alone (ViSpGAN w/o $\mathcal{L}_{fm}$, $D_{insp}$), and as expected, its performance is similar to 3D StarGAN both in terms of accuracy and FID score. 

Finally for comparison, we also perform the same evaluation protocol using the ViSpGAN model trained and tested on the entire dataset of 500 words. We report the results in Table \ref{table:table1}. We observe that the FID score has been greatly improved from training the model on the entire LRW dataset (which also corresponds to better visual quality of the generated videos and will be demonstrated in the next section on qualitative analysis). The accuracy stays high even for $500$ word classification. 


\begin{figure*}[t]
\begin{center}
   \includegraphics[width=0.89\linewidth]{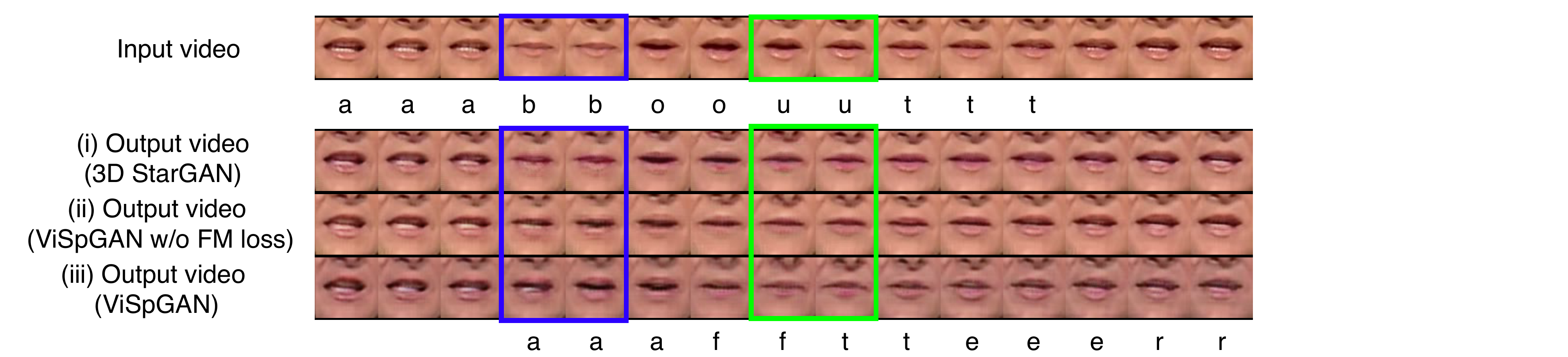}
\end{center}
\caption{Generated videos conditioned on the same video input from the test set of LRW, using: (i) 3D StarGAN, (ii) ViSpGAN without feature matching, and (iii) ViSpGAN when trained on 50 words. The original word is \textit{"about"} and the target word is \textit{"after"}. The `blue' box shows how the models generate letter \textit{`a'} (slight mouth opening in the 3D StarGAN output, pronounced lip movement for both ViSpGAN models). In the `green' box ViSpGAN model produces more conspicuous lip movements for the letter combination \textit{`ft'} in comparison with the other two models.}
\label{fig:fig5}
\end{figure*}
\begin{figure*}[h]
\centering
\begin{subfigure}[b]{1.0\textwidth}
   \includegraphics[width=1\linewidth]{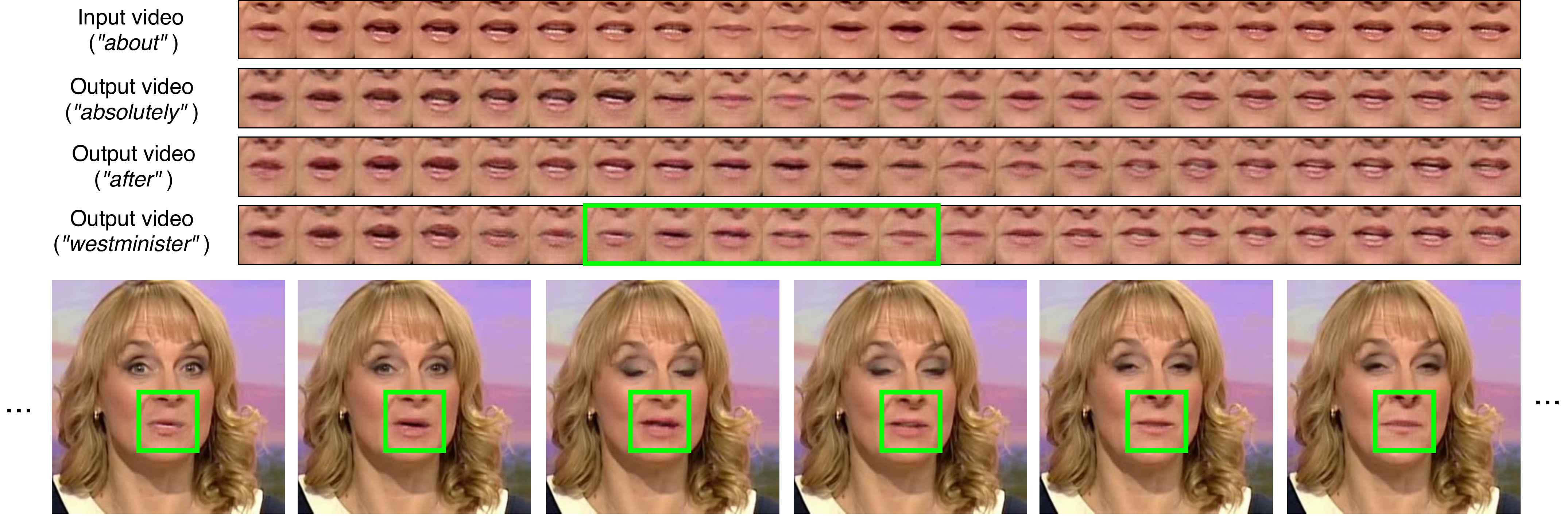}
   \caption{}
   \label{fig:fig4a} 
\end{subfigure}

\begin{subfigure}[b]{1.0\textwidth}
   \includegraphics[width=1\linewidth]{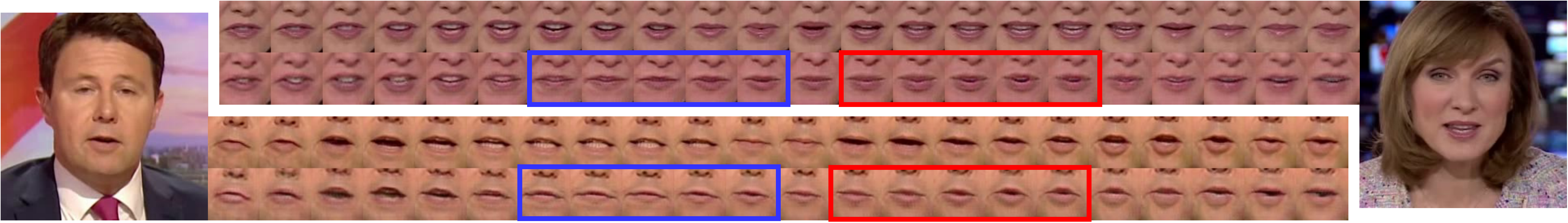}
   \caption{}
   \label{fig:fig4b}
\end{subfigure}

\caption[]{Videos generated with our ViSpGAN model trained on 500 words. (a) Given an input video with the word \textit{"about"} and three target words \textit{"absolutely"}, \textit{"after"} and \textit{"westminister"}, the generator synthesizes three corresponding videos. 
(b) Given two input videos and the same target word \textit{"football"}, the generator produces two fake videos with consistent lip movements of the target word. For each video pair, the first one is the input to the generator and the second one is the output. In the `blue' colored box, the uttering corresponds to \textit{"foot"}, while the `red' colored box corresponds to \textit{"ball"}.}
\end{figure*}

\subsection{Qualitative analysis}
We start with the qualitative comparison of the fake samples generated by the 3D StarGAN and the ViSpGAN models, when trained on the LRW subset of 50 words. In Fig. \ref{fig:fig5}, we show an example of the generated videos given an input video with uttering \emph{"about"} and a target word \emph{"after"}. As we can see, the video output of 3D StarGAN closely follows the input video. We can notice a slight `mouth opening' in the blue box, a behaviour that corresponds to the uttering of letter \textit{`a'}. The ViSpGAN models (with and w/o feature matching) create a much more plausible lip movement for \textit{`a'} at the same frames. Furthermore, in the green box, we observe that ViSpGAN produces more plausible and conspicuous lip movements for the letter combination \textit{`ft'}, in comparison with 3D StarGAN and ViSpGAN w/o feature matching. We attribute this to the fact that the character labels along with the character inspector force the generator to learn visual features that correspond directly to letters. 

Next we demonstrate the ability of our model to synthesize different target words \emph{"absolutely"}, \emph{"after"} and \emph{"westminister"}, given the same video input. 
For this task, we use the ViSpGAN model trained on the entire LRW dataset.  As can be seen in Fig. \ref{fig:fig4a}, the generator has learned to modify the input video according to the target word and produces three different word utterings. Changes are most pronounced in the middle of the video, where the actual word resides. 
Finally, in Fig. \ref{fig:fig4b}, we show that even when the generator is given inputs which are dissimilar to each other, it successfully synthesizes videos with consistent lip movements \emph{for the same target word} (\emph{"football"}). We use Poisson Editing \cite{Perez} to blend the generated mouth region back into the original video with the speaker. More results including videos are in the supplementary material.

\section{Conclusions}
We presented ViSpGAN, a GAN-based framework for video-to-video translation, for the task of visual speech synthesis. We conditioned video generation on the video of a speaker and a target word encoded as a sequence of characters. We showed that synthetic samples generated by the proposed ViSpGAN model contain strong visual speech features, similar to the real videos, when assessed by an auxiliary lipreading network. 
The advantage of the proposed model is that in principle, it can produce outputs of any length. By applying the generator sequentially on a long video we could produce multiple words based on their characters. 
This could be a way of generating videos of full sentences and it is an interesting future direction to explore. 
\section*{Acknowledgments}
\noindent 
VS is supported by the Imperial College Research Fellowship.
We gratefully acknowledge NVIDIA for GPU donation and Amazon for AWS Cloud Credits.
{\small
\bibliographystyle{ieee}
\bibliography{egbib}
}

\end{document}